\documentclass[conference]{IEEEtran}
\IEEEoverridecommandlockouts
\usepackage{cite}
\usepackage{multirow}
\usepackage{graphicx}
\usepackage{float}
\usepackage[table,xcdraw]{xcolor}
\usepackage{amsmath,amssymb,amsfonts}
\usepackage{textcomp}
\usepackage{hyperref}
\usepackage{algorithm}
\usepackage{algpseudocode}

\def\BibTeX{{\rm B\kern-.05 em{\sc i\kern-.025em b}\kern-.08em
    T\kern-.1667em\lower.7ex\hbox{E}\kern-.125emX}}
\begin{document}

\title{Curiosity-driven Intuitive Physics Learning\\
}

\author{\IEEEauthorblockN{Tejas Gaikwad}
\IEEEauthorblockA{\textit{Dept. of Computer Science and Engineering} \\
\textit{Indian Institute of Technology Jodhpur}\\
Rajasthan, India \\
gaikwad.2@iitj.ac.in}
\and
\IEEEauthorblockN{Romi Banerjee}
\IEEEauthorblockA{\textit{Dept. of Computer Science and Engineering} \\
\textit{Indian Institute of Technology Jodhpur}\\
Rajasthan, India \\
romibanerjee@iitj.ac.in}
}

\maketitle

\begin{abstract}
Biological infants are naturally curious and try to comprehend their physical surroundings by interacting, in myriad multisensory ways, with different objects - primarily macroscopic solid objects - around them. Through their various interactions, they build hypotheses and predictions, and eventually learn, infer and understand the nature of the physical characteristics and behavior of these objects. Inspired thus, we propose a model for curiosity-driven learning and inference for real-world AI agents. This model is based on the arousal of curiosity, deriving from observations along discontinuities in the fundamental macroscopic solid-body physics parameters, i.e., shape constancy, spatial-temporal continuity, and object permanence. We use the term 'body-budget' to represent the perceived fundamental properties of solid objects. The model aims to support the emulation of learning from scratch followed by substantiation through experience, irrespective of domain, in real-world AI agents. 
\end{abstract}


\section{Introduction}
How do we learn? How do we choose what to learn? How do we
continually store and update knowledge? When do we understand that we have understood?... It's been centuries that philosophers and scientists have been mulling over such questions, and these have- and continue to- influence research across disciplines; the fields of Artificial Intelligence and Machine Learning \cite{b1} being at the pinnacle of such endeavours.
Despite present-day machine learning advancements, it is difficult for a machine to execute a task that a human naturally performs. There is yet a huge gap between a machine's ability to observe, encode experiences and formulate reasons.\\
Here we present an approach that can potentially be used to design machines that can 'understand' the physical world around them. Our model draws inspiration from the way infants learn \cite{b2}. It begins with parsing the surroundings by sensing all that is in the physical space around them, interacting with the objects to understand the dynamics, experimenting with them, and deriving hypotheses and predictions, and eventually reasoning on the body budget parameters (shape constancy, spatial-temporal continuity, object permanence). Sensing refers to touch, smell, hearing, smell, and taste. Physical space is the materialistic world around infant.These basic senses help humans identify components in their physical space and their behavior,  followed by updating knowledge gathered on the objects through reasoning. Whenever an unexpected event happens, curiosity gets triggered, which contributes to the urge to 'know or learn more' about such an event, further supported with valid reasons (with as much confidence as is possible).\\
The way reinforcement learning works is broadly aligned with the biological infants learning about the physical space. They perceive,
they act, develop the knowledge about the physical space, build the
parameters about it, interlink the knowledge with the parameters to have a virtual simulation of the physical space around them. The infants learn a lot by
observing the actions, states, results and then try to imitate similar actions.\\
In our work, we have focused on the visual sense to demonstrate the proposed method for curiosity-driven learning. We have considered three fundamental solid body physics properties, essential for the comprehension of object-dynamics, i.e., Spatial-temporal continuity, Object Permanence, and Shape constancy, as discussed in the paper \cite{b4}.
\section{Related Work}
By focusing on learning as a function of inferences on fundamental properties of solid objects, the designed mechanism is envisioned to serve as a substrate for generalized learning and transfer learning. Some design inspirations are:
\begin{itemize}
\item Nguyen et al. 2020\cite{b4} discuss the state-of-the-art method to learn intuitive physics with the help of measuring the element of surprise and explaining it. Surprises are explained by a function of motion-based surprise
and appearance-based surprise parameters. The evaluation metrics used were Absolute Error Rate and Relative Error Rate for parameters of Object Permanence,
Shape Constancy, and Spatial-temporal Continuity.
\item Pathak et al. (2017)\cite{b5} used curiosity to explain the need to explore the environment and discover novel states. Curiosity is considered
an intrinsic reward that enables the agent to explore the environment more.
The method empowers an agent to learn generalizable skills even
in the absence of an explicit goal.
\item Zhou et al. (2020) \cite{b6} present 'meta-imitation learning' as a mechanism of learning from few demonstrations and trying to imitate them through trial-n-error.
\item Pal et al.(2013) \cite{b7} describe a way of representing a knowledge using Zadeh's Z-numbers \cite{b12}. A Z-number is of the form \(<X, A, B>\), where 'X' is the subject, 'A' being predicate, a value from a set of possible values for X, and B is the confidence in 'X'='A'. The concept can be visually envisaged as the building blocks of a knowledge graph where 'X', 'A' represent the graph nodes, and B represents the edge weights. The edge weights are the function of reinforcement. 'X' and 'A' are the function of an experience.
\item Chitnis et at. (2019)\cite{b9} describes a technique where an agent learns to pick and operate on objects of different shapes and sizes. They trained the agent in a multitask setting to learn different distribution of object shape and sizes. Tanwani et al.(2020) \cite{b11} describes a semi-supervised approach, named Motion2vec, towards acquiring manipulation skills shown in surgical videos for surgical suturing. The main idea behind this approach is learning  invariant representations from surgical videos for tasks of action segmentation and pose intimation.
\item Meier et al. (2017) \cite{b10} proposed an approach for learning to learn while learning. Obtained results have proven fastening initial learning and faster convergence on subsequent tasks.  
\end{itemize}



\section{Methodology and System Design}
We propose a method for curiosity-driven learning of intuitive physics through interactions with everyday objects. We have considered object's body-budget parameters (shape constancy, spatial-temporal continuity, and object permanence) to parse and understand the physical world around. As stated in the article \cite{b4}, these parameters come
handy while parsing natural physical objects-object interactions. 
All these parameters are represented in the form of a linear plot w.r.t. time / frame-sequence. These plots demonstrate the behavior of the object in a given event (from the data-set) as a line plot. These plots are then examined to identify discontinuity in the object responses if any, and further trigger the curiosity component if discontinuity is observed. 
\subsection{Dataset}
\begin{figure}[h!]
\centering \includegraphics[width = 0.48\textwidth]{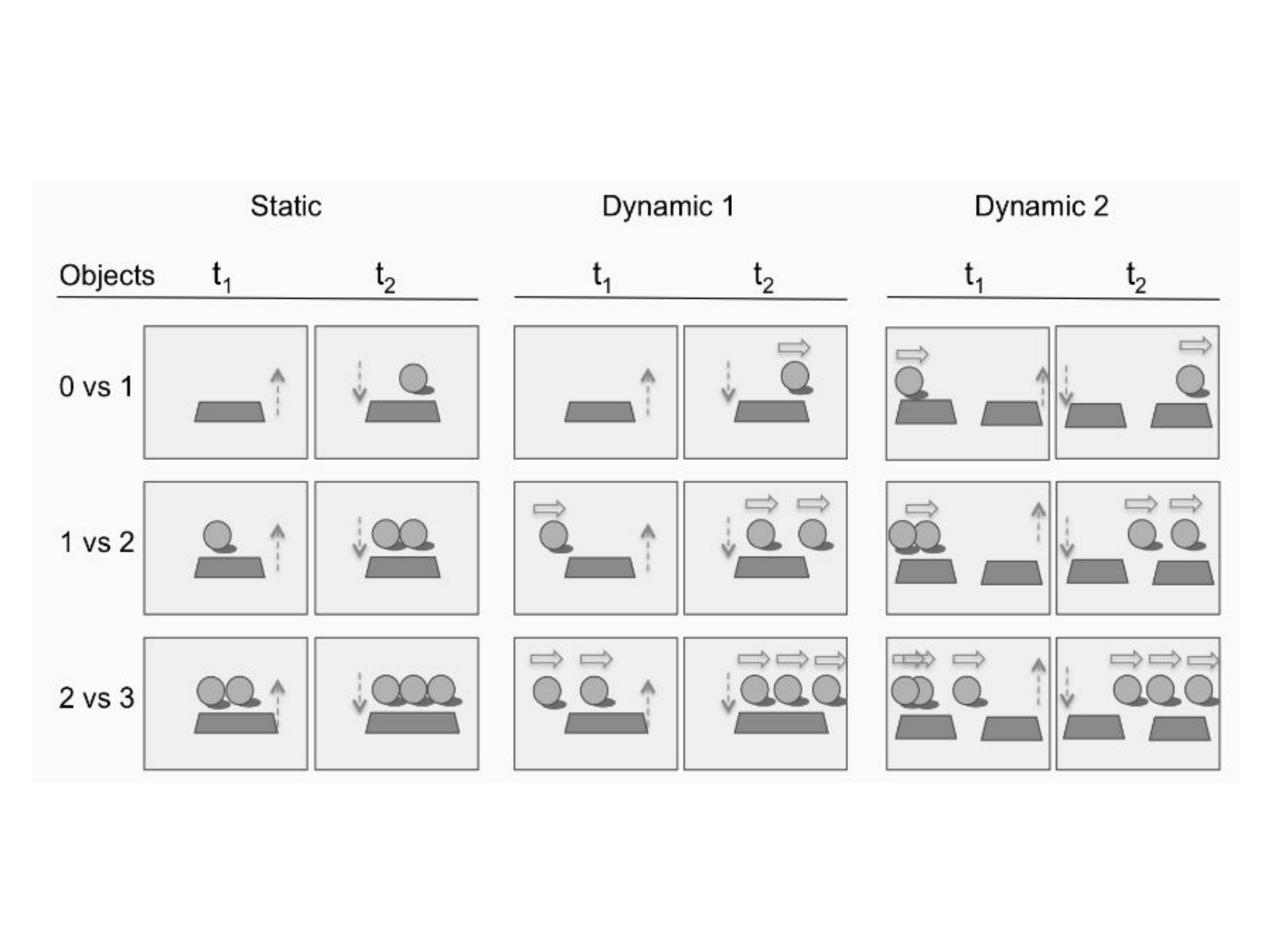}
  \caption{Dataset Sample Event Representation. In above image, we have 2 object types, a sphere and a wall constituting 9 different scenarios as shown. Diagram taken from Intuitive Physics Dataset Website}
  \label{Sample Event1}
\end{figure}

Dataset used for learning and training of our agent is the Intuitive Physics Dataset \cite{b13}. This dataset constitutes a set of possible and impossible events presented through animations. Every event in the dataset has some objects, which can be any or all of a cube, cone, and a sphere. These events may or may not contain an occluder which is a wall in our case. These events can be further classified as static or dynamic based on the object's movement in the event. The sample event of possible and impossible with occluder as a wall is shown in Fig.\ref{Sample Event1}
\subsection{System Architecture}
\begin{figure}[h!]
  \centering \includegraphics[width = 0.48\textwidth]{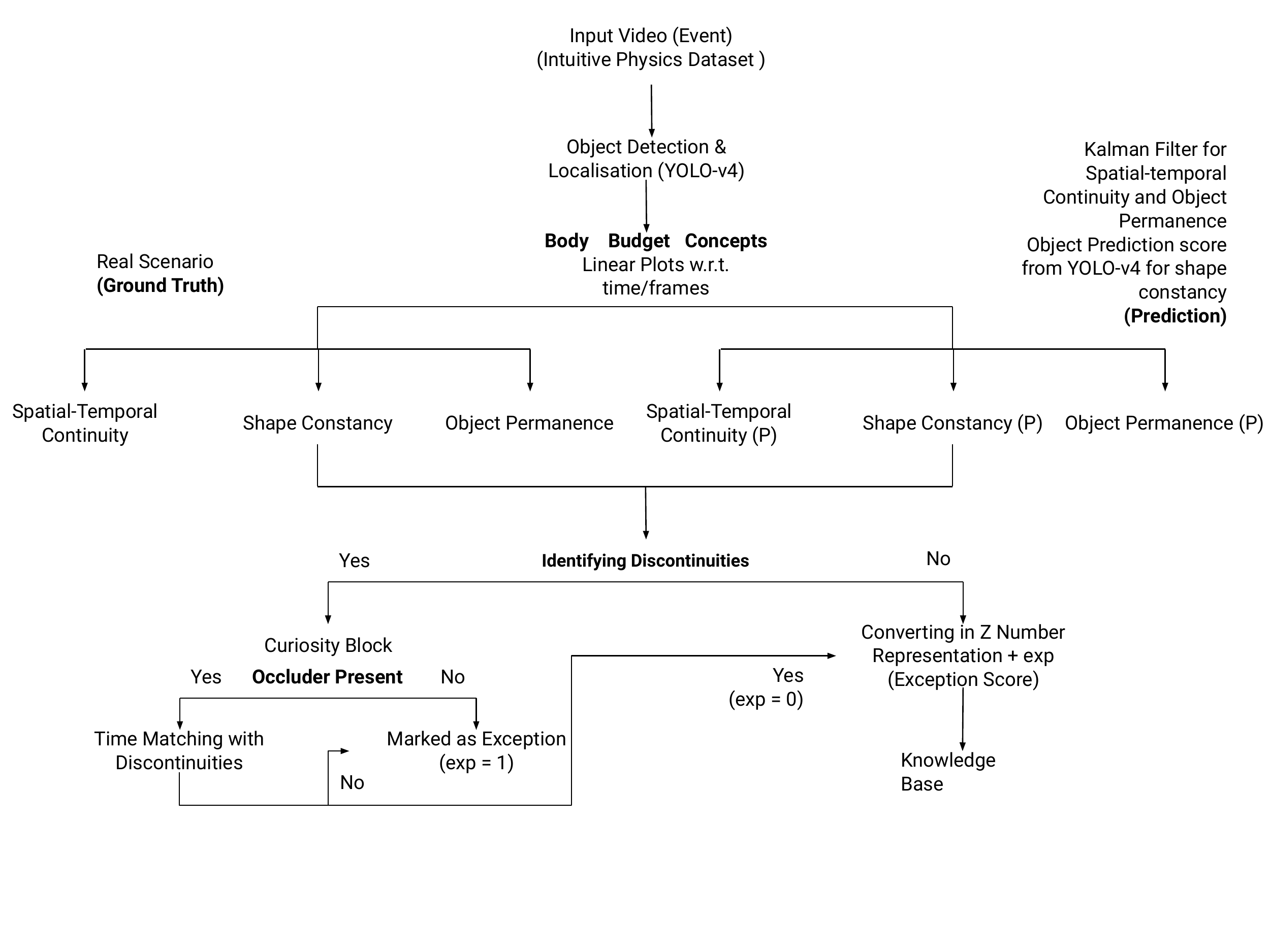}
  \caption{System Architecture}
  \label{proposed solution}
\end{figure}

\begin{figure}[h!]
  \centering \includegraphics[width = 0.4\textwidth]{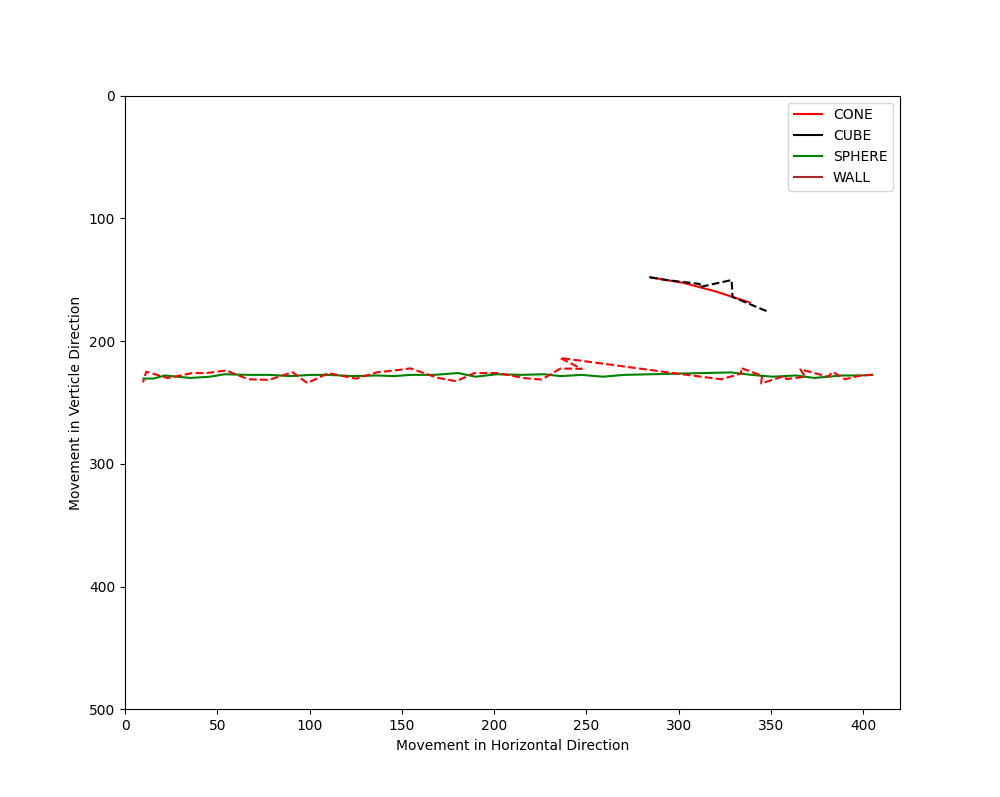}
  \caption{Body Budget parameters Linear Plot. The continuous line represent the actual path travelled and the dotted represents the predicted path. Different colors are used for different object class}
  \label{Sample Event}
\end{figure}

The proposed solution is shown in Fig \ref{proposed solution}. The video input is
given to YOLO-v4 model for object detection and localization\cite{b8}. The localized object information is used to obtain linear plots for the body budget parameters. These linear plots are then used to analyze discontinuities in the responses as shown in Fig \ref{Sample Event}. These discontinuities are detected by comparing the obtained responses for parameters of body budget with the predicted one. The predictions for the body budget parameters are made using Kalman Filter. Such discontinuities trigger the curiosity block in the agent's mind. The curiosity block makes the agent revisit the previous event and check if there is any external factor affecting the expected responses of the body budget, factors like the presence of an occluder. If occluder is detected for the same time when discontinuities are observed, then the agent accepts this as the reason for the object behavior, and matches the decision with the ground truth which is already provided with the dataset. If the responses are matched, the agent proceeds further for a new event, else it is labeled as an exceptional case. This exception case is still stored in the knowledge base of the agent, which further seeks similar responses to gain confidence about that event. If the agent cannot find reasoning for the discontinuities, it is tagged as an impossible event and further matched with the ground truth. If ground truth matches with the labeling by the agent, then it proceeds further for a new event else added as an exception. The exception added is further generalized if the number of events gets the same responses for the same state and is also labeled as a reasoning label, the same way we humans accept some facts which we might not aware of the reasoning yet but accepts the existence of it. 

\subsection{Knowledge Update and Inference}
\begin{itemize}
    \item A generated score for each body budget parameter represents unique characteristics about the object in the given environment. 
\item We consider a hypothetical space that we term latent space. The latent space is similar to plotting data points in a 2D plane and classifying these data points using clustering. Latent space is divided into three spaces that represents classes. 
\item We plot the scores obtained for an unknown object in this latent space. The closer the calculated data point, the more confident the agent will be to classify the unknown object as a specific class. 
The score generated is a function of parameters of body budget multiplied with constants $(\alpha, \beta,\gamma)$ with values ranging from 0 to 1, determining the priority of a particular concept. Priority is decided on the basis of the presence of the occluder. 
\item These scores then are represented in the form of a Z-number. A Z-number of the form \(<X, A, B>\) in which ‘X’ represents a class, ‘A’ represents the score calculated as shown in equation (1), and ‘B’ represents confidence which is calculated as shown in equation (4) and inference is shown in equation \eqref{Inferencing Z}

\end{itemize}
\subsection{Curiosity Driven Learning}
\begin{itemize}
    \item With a discontinuity in the linear plot of the body budget, the curiosity block gets activated. 
\item On activation, the agent revisits the event and tries to find out the reasoning for the discontinuity as shown in Fig. \ref{proposed solution}
\item If no reasoning found, it will add exceptional value to the reasoning and keep the information as per ground truth. 
\item With further past experience, this exception will be removed if multiple similar experiences are observed.  

\end{itemize}

\begin{figure}[h!]
  \centering \includegraphics[width = 0.48\textwidth]{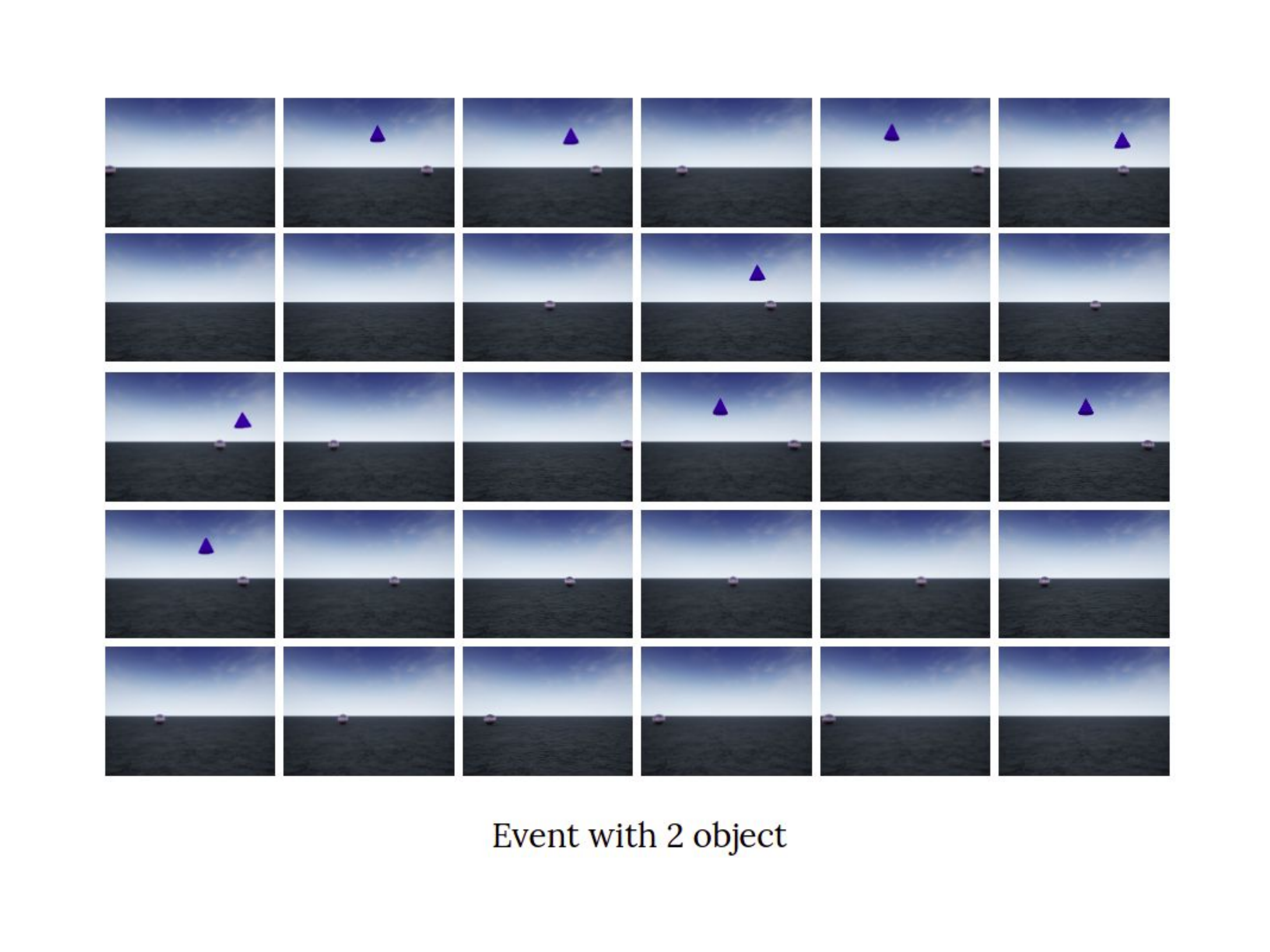}
  \caption{Example Event. This is the actual dataset sample event whose body budget concept's Linear plot is shown in Fig. \ref{Sample Event}. This event has 2 different objects, a sphere and a cone moving from one end to another. Sphere follows the normal path and does not have any surprise factor whereas cone suddenly appears in one time frame, follows a continuous path and then disappears which give rise to curiosity Riochet et al.\cite{b13} }
  \label{example Event}
\end{figure}
\subsection{Equations}
As discussed in Section III.B, Z-numbers constitutes of \(<X,A,B>\).
For an unknown object, 'X' will be null initially. 'A' being the score of the object w.r.t. the body budget parameters is shown in equation \eqref{A}. Score generated for object permanence is given in equation \eqref{sop}.
\begin{equation}
A = \alpha*S_{op} + \beta*S_{sc} + \gamma*S_{stc}
\label{A}
\end{equation}
\(S_{op}\) is Score for Object permanence, \(S_{sc}\) is Score for Shape Constancy, \(S_{stc}\) is Score for Spatial Temporal Continuity.\(S_{sc}\) is obtained from YOLO-v4 object prediction scores.
\begin{equation}
 S_{op} = \frac{\sum_{n=1}^{N} P_{Yolo}(O(x_{t}))*Impact Value }{1000} 
\label{sop}
\end{equation}
 \(P_{Yolo}(O(x_{t}))\) is the prediction score of the detected object. Impact Value is the unique value specific to a class. For our consideration, we have assigned constant value of 10 for sphere, value of 100 for cone and a value of 1000 for cube. As we are doing summation, we will be having these value in terms of 100's etc. giving unique score for a given class. 'N' represents total number of frames. The range for \(S_{op}\) will be different for different class having unique range for each class. \(S_{sc}\) is normalised euclidean distance between object's images. 
 \begin{equation}
S_{stc} = 1 - \frac{N - Count\ O(x_{t})}{N}
\label{stc}
\end{equation}
Here, Count O(\(x_{t}\)) is the count of frames in which the object was detected.
\begin{equation}
B_{c} =\frac{\frac{\begin{vmatrix}
A_{avg_c} - A_{u} 
\end{vmatrix}}{A_{avg_c}}}{\sum_{1}^{N} \ \frac{\begin{vmatrix}
A_{avg_c} - A_{u} 
\end{vmatrix}}{A_{avg_c}}}, 
B\epsilon\ [0,1], C \epsilon \begin{Bmatrix}
Cube, Sphere, Cone
\end{Bmatrix}
\label{B Value}
\end{equation}
 Here, \(A_{avg_c}\) is average 'A' score for a particular class, \(A_{u}\) is 'A' value of unknown class. Based on this, 'Z' will be calculated as in equation \eqref{Inferencing Z}
\begin{equation}
Z = C, for\ C\ with\ min\ (B_{c}) \label{Inferencing Z}
\end{equation}
\section{Experiment}
For the experiment described in Table 1, we considered the ground truth as 'sphere' and obtained the following values: N = 90, \(P_{yolo}(O(x(t))\) = 0.6 (for sphere) and 0.17(for cone). As this does not involve any occluder, $\alpha=\beta=\gamma=0.33$, Count\ \(O(x_{t})\) for sphere = 48, for cone = 0. 
\begin{table}[hbt!]
\resizebox{0.48\textwidth}{!}{%
\begin{tabular}{|l|l|l|l|l|l|l|l|c|}
\hline
\begin{tabular}[c]{@{}l@{}}Object\\ Ground\\ Truth\end{tabular} &
  \begin{tabular}[c]{@{}l@{}}Object\\ Assumed\end{tabular} &
  $S_{op}$ &
  $S_{stc}$ &
  $S_{sc}$ &
  $A$ &
  $A_{avg}$ &
  $B$ &
  \multicolumn{1}{l|}{\begin{tabular}[c]{@{}l@{}}Inference\\ (Z)\end{tabular}} \\ \hline
\multirow{2}{*}{\begin{tabular}[c]{@{}l@{}}Ground\\ Truth\end{tabular}} &
  Sphere &
  0.288 &
  0.53 &
  0.68 &
  0.494 &
  1.57 &
  1.076 &
  \multirow{2}{*}{Sphere} \\ \cline{2-8}
 &
  Object &
  8.16 &
  0 &
  0.27 &
  2.78 &
  17.88 &
  15.08 &
   \\ \hline
\end{tabular}%
}\\
\caption{\label{tab:table-name}Experiment Results}
\end{table}
\begin{algorithm}
\caption{Algorithm}
\begin{algorithmic}[1]
\State $\textrm{CD()}\ as\ \textrm{Curiosity\ Driven\ Function}$
\State $x_{t} \gets \textrm{Frame\ at\ Time}\ t$
\State $x_{t-1} \gets \textrm{Frame\ at\ Time}\ t-1$
\State $O(x_{t}) \gets \textrm{Objects\ located\ in}\ x_{t}\ \textrm{frame}$
\State $G(x_{t}) \gets \textrm{Ground\ truth\ of\ objects}$ 
\State $EoF \gets \textrm{End\ of\ frame}$\\
$O(x_{t}), G(x_{t}) \in \{C_1, C_2, C_3, C_4\}$
\State $C_1,\ C_2,\ C_3,\ C_4\ are\ classes$

\While {EoF\ != True}
    \If {O$(x_{t}) \in \{C_1, C_2, C_3\}$}
        \If {O($x_{t}$) == O$(x_{t-1})$}
            \State $FlagEvent \gets \textrm{Possible}$
            \State $count\ O(x_{t}) += 1$
        \Else 
            \State $FlagEvent = \textrm{CD()}$
        \EndIf
    \ElsIf {$O{x_{t-1}} \notin \{C_1, C_2, C_3\}$}
        \State $FlagEvent \gets \textrm{CD()}$
    \EndIf
\EndWhile

\Procedure{CD}{ }
    \While {EoF != True}
        \If {O$(x_{t}) \in C_4$}
            \State $countwall$ += 1
        \EndIf
        \If{$countwall$ in range $(0.7 \times \textrm{count\ O}(x_{t})+\textrm{count\ O}(x_t), \textrm{count\ O}(x_t)))$}
            \State $FlagEvent \gets \textrm{Possible}$
        \Else
            \State $FlagEvent \gets \textrm{Not Possible}$
        \EndIf
    \EndWhile\\
    \Return $FlagEvent$
\EndProcedure

\end{algorithmic}
\end{algorithm}


\section{Conclusion and Discussions}
To the best of our knowledge, the proposed idea of curiosity-driven learning mechanism is a novel technique. It shows promise in applications / areas that require learning from scratch and learning from learning. Some directions for future work are as follows,
\begin{itemize}
\item Incorporation of a knowledge graph towards never-ending learning and Z-number based in-policy reinforcement and transfer learning.
\item Knowledge-clustering and generation of concept-clouds and 'subjective' experiences.
\item  Robots learning from scratch, and eventually from experiences, in any domain. 
\end{itemize}

\end{document}